# 3D 신경망과 어텐션 메커니즘을 이용한 행동 인식

러비나 스레스터[*1], 시카 두베[*2], 파루크 올리모브[3], 무하마드 아심 라피크[2], 전문구[2]
[1] 데이터 팀, (주)초위스
[2] 전저전기컴퓨터공학, 광주과학기술원
[3] 위협 인텔리전스 팀, 모니터랩
labina.shr@gm.gist.ac.kr, shikha.d@gm.gist.ac.kr, olimov.farrukh@gist.ac.kr,
aasimrafique@gist.ac.kr, mgjeon@gist.ac.kr

## 3D Convolutional with Attention for Action Recognition

Labina Shrestha[*1], Shikha Dubey[*2], Farrukh Olimov[3], Muhammad Aasim Rafique[2], Moongu Jeon[2]
[1] Data Team, Chowis, Gyeongi-do, South Korea,
[2] School of Electrical Engineering and Computer Science,
Gwangju Institute of Science and Technology, Gwangju, South Korea,
[3] Threat Intelligence Team, Monitorapp, Seoul, South Korea

Abstract

Human action recognition is one of the challenging tasks in computer vision. The current action recognition methods use computationally expensive models for learning spatio-temporal dependencies of the action. Models utilizing RGB channels and optical flow separately, models using a two-stream fusion technique, and models consisting of both convolutional neural network (CNN) and long-short term memory (LSTM) network are few examples of such complex models. Moreover, fine-tuning such complex models is computationally expensive as well. This paper proposes a deep neural network architecture for learning such dependencies consisting of a 3D convolutional layer, fully connected (FC) layers, and attention layer, which is simpler to implement and gives a competitive performance on the UCF-101 dataset. The proposed method first learns spatial and temporal features of actions through 3D-CNN, and then the attention mechanism helps the model to locate attention to essential features for recognition.

## 1. INTRODUCTION

Action recognition has become a prevalent topic within the computer vision field in the past few years, especially after implementing deep learning techniques in this domain. Humans' action recognition in video is of interest for applications such as visual surveillance [7], behavior monitoring, human-computer interaction, and video understanding [6]. Similar to the other areas of computer vision, most of the action recognition algorithms are based on convolutional neural networks (CNNs). Before introducing 3D-CNN, CNN architectures were generally used for recognizing actions in a static image as their performance was based on spatial analysis. However, unlike image recognition, action recognition works on video frames, which means that we also need to learn temporal information. Due to the temporal dimension in video datasets, the amount of data to be processed is high, which makes the model more complex and recognition more challenging.

---

* Equal Contribution

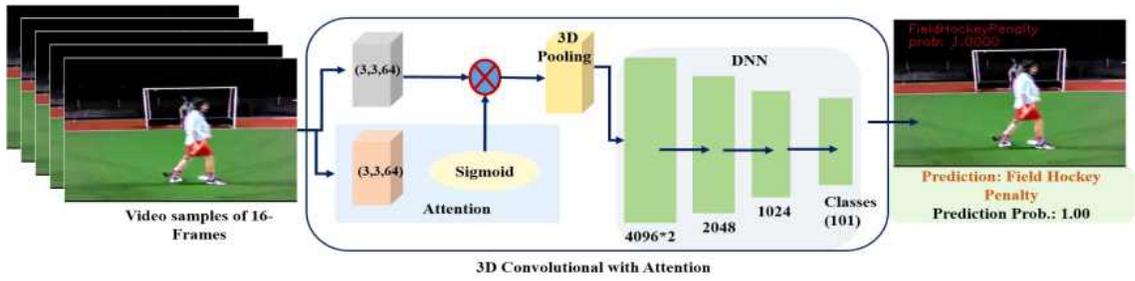

Fig.1.. The proposed architecture for action recognition.

To analyze the temporal information in videos, many models have been introduced. Two-stream convolutional network [1] separately extracts spatial and temporal features from RGB and stacked optical-flow images. Long-term recurrent convolutional network (LRCN) [2] learns spatial features from each frame and then learns temporal features using recurrent neural networks (RNN). These methods use spatiotemporal features separately in different stages. However, learning spatio-temporal features simultaneously from videos will be more effective for action recognition. Using 3D-CNN [3] also seems like a natural approach as it can learn spatio-temporal information using a single model.

To improve the performance of 3D-CNN, a two-stream inflated 3D ConvNet [3] was introduced where 3D ConvNets can directly learn about temporal patterns from an RGB stream. However, to improve the performance, they included an optical-flow stream as well. The previously proposed methods were expensive to train as they needed extra streams or parameters to increase the performance.

In this paper, we propose a simple yet efficient method for action recognition in video. The proposed method utilizes a 3D-CNN to extract spatio-temporal information and attention mechanism to give attention to essential features for recognition and likewise learns spatio-temporal dependencies. The details and experimentations of the proposed model are given in the following sections.

## 2. The Proposed Action Recognition Model

The proposed model (Figure 1)is composed of four components: input preprocessing, 3D-CNN, attention gate & average pooling, and dense neural network (DNN). As 3D-CNN uses a 5D tensor (batch size, channel, height, width, depth) as input, we change the number of frames accordingly. We use RGB channels with 16 number of frames as input tensors. The 3D-CNN of channel 64, followed by the ReLU activation function, extracts spatial and temporal features from input frames. These extracted features are elementwise multiplied with the output of the attention layers to get the essential features. The attention layer consists of sigmoid gating of 3D-CNN extracted features, which helps the model to focus on the more essential features for recognizing the action. After the attention layer, a 3D average pooling is utilized. Then these extracted features are passed through DNN layers followed by ReLU activation with a dropout rate of 0.25 and predicts the action for the given video sample of 16-frames.

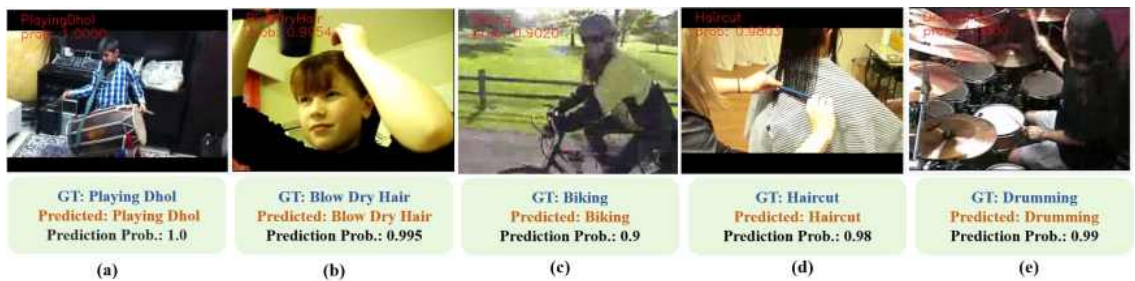

Fig..2. Qualitative analysis of the proposed method on UCF-101 dataset. In (a)-(e), model predicts all the actions correctly with higher probability. GT stands for ground truth.

## 3. Implementation and Evaluation
### 3.1 Dataset

We trained our model on the UCF101 dataset [5] with 13320 videos from 101 action categories. The videos in 101 action categories are grouped into 25 groups, where each group can consist of 4-7 videos of an action. The datasets were split into 8460 training sets, 2156 validation sets, and 2701 test sets.

### 3.2 Implementation Details

The model is build using PyTorch 1.8 on a machine with NVidia GeForce RTX 2080 and RAM 8GB. The model was trained for 50 epochs with an initial learning rate of 1e-4 and weight decay with the rate of 1e-4 after 4 steps followed by Adam optimizer. Data augmentation is known to be of crucial importance for the performance of deep learning architecture. We used spatially and temporally random cropping by resizing the video frame size to 128*171 pixels, then randomly cropping 112 x 112 patch. For temporal cropping, we randomly selected the time index for temporal jittering. The spatial crop is performed on the entire frame, so each frame is cropped on the exact location. The temporal jitter takes place via the selection of consecutive frames. We also applied random horizontal flips consistently for each video during training.

### 3.3 Evaluation

We have evaluated the model using a soft-max classifier for the classes of UCF-101. An additional mechanism like Fine-tuning, ensemble networks were not used. Moreover, pre-trained networks were also not used. We evaluated the test/val data of UCF-101. The evaluation results are presented in Table1, Figures 2 and 3. Table 1 shows that our proposed model, 3D-CNN with attention, outperforms 3D-CNN along with other mentioned models. Figure 2. shows a qualitative analysis of our proposed model on the UCF-101 test dataset. Figure 3 shows the comparison between 3D-CNN and our model using the accuracy and loss graph for each epoch on train, val, and test datasets.

## 4. Conclusion and Future work

This paper has proposed a method, 3D convolution with attention for the action recognition task, which outperforms the other methods on the UCF-101 dataset without using additional fusion streams and without fine-tuning with large-scale datasets. Our future works will include further training using large-scale datasets or mega-scale datasets and deep CNN networks to enhance spatiotemporal 3D CNNs.

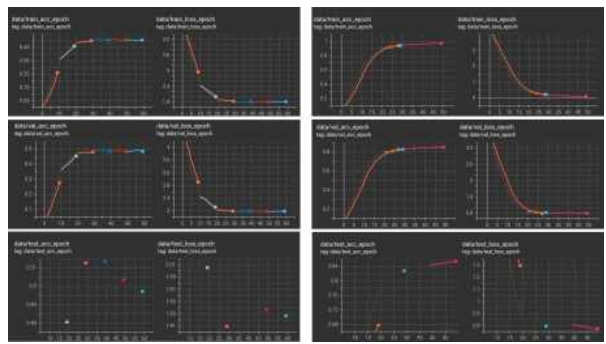

Fig.3. Comparison between 3D-CNN (Column 1) and 3D-CNN+Attention (Ours) (Column 2) models. The first, second, and third row represent train, val, and test evaluation respectively.

Table 1 Comparison of our proposed model with other models on UCF-101 test and val dataset.

| Model | UCF-101 |
|---|---|
| CNN+LSTM[2] | 68.20 |
| 3D-ConvNet [3] | 52.8 |
| 3D-Fused [4] | 69.5 |
| 3D-ConvNet + Attention (Val) (Ours) | 81.1 |
| 3D-ConvNet + Attention (Test) (Ours) | 82.5 |

## Acknowledgement

This work was supported by Institute of Information communications Technology Planning Evaluation (IITP) grant funded by the Korea Government (MSIT) (No. 2014-3-00077, Development of Global Multi-target Tracking and Event Prediction Techniques Based on Real-time Large-Scale Video Analysis) and Korea Creative Content Agency (KOCCA) in the culture Technology (CT) Research Development Program (R2020060002) 2021.